\documentclass[accepted]{uai2025} 
                        
\usepackage[american]{babel}
\usepackage{amsmath}
\usepackage{algorithm}
\usepackage{algorithmic}
\usepackage{booktabs}
\usepackage{multirow}
\usepackage{tabularx}
\usepackage{array}
\usepackage{makecell}
\usepackage{float}

\usepackage{natbib} 
\usepackage{mathtools} 
\usepackage{booktabs} 
\usepackage{tikz} 

\title{RDI: An adversarial robustness evaluation metric for deep neural networks based on model statistical features}

\author[1,3]{Jialei Song}
\author[1,3]{Xingquan Zuo$^*$}
\author[1,3]{Feiyang Wang}
\author[1,3]{Hai Huang}
\author[2,3]{Tianle Zhang}
\affil[1]{%
    Shool of Computer Science, Beijing University of Posts and Telecommunications, Beijing, China
}
\affil[2]{%
    Shool of Cyberspace Security, Beijing University of Posts and Telecommunications, Beijing, China
}
\affil[3]{%
    Key Laboratory of Trustworthy Distributed Computing and Service, Ministry of Education, Beijing, China
  }
  
\begin{document}
\maketitle

\begingroup
\renewcommand\thefootnote{*}
\footnotetext{Corresponding author}
\endgroup

\begin{abstract}
  Deep neural networks (DNNs) are highly susceptible to adversarial samples, raising concerns about their reliability in safety-critical tasks. Currently, methods of evaluating adversarial robustness are primarily categorized into attack-based and certified robustness evaluation approaches. The former not only relies on specific attack algorithms but also is highly time-consuming, while the latter due to its analytical nature, is typically difficult to implement for large and complex models. A few studies evaluate model robustness based on the model's decision boundary, but they suffer from low evaluation accuracy. To address the aforementioned issues, we propose a novel adversarial robustness evaluation metric, Robustness Difference Index (RDI), which is based on model statistical features. RDI draws inspiration from clustering evaluation by analyzing the intra-class and inter-class distances of feature vectors separated by the decision boundary to quantify model robustness. It is attack-independent and has high computational efficiency. Experiments show that, RDI demonstrates a stronger correlation with the gold-standard adversarial robustness metric of attack success rate (ASR). The average computation time of RDI is only 1/30 of the evaluation method based on the PGD attack. Our open-source code is available at: \url{https://github.com/BUPTAIOC/RDI}.

\end{abstract}

\section{Introduction}\label{sec:intro}
With the rapid advancement of deep learning, deep neural networks (DNNs) have achieved unprecedented success in fields such as computer vision \cite{ruiz2023dreambooth}, natural language processing \cite{alberts2023large}, and multimodal tasks \cite{liupatch}.
However, recent studies have shown that even DNNs that perform excellently on standard benchmark datasets may experience significant performance degradation when facing with small perturbations \cite{goodfellow2014explaining}. These minor disturbances, especially adversarial examples generated by advanced attack algorithms, reveal the vulnerabilities of DNNs in terms of adversarial robustness. Consequently, effectively assessing the adversarial robustness of DNNs has become one of the current research hotspots.

Research on adversarial robustness assessment has primarily focused on two directions: attack-based \cite{moosavi2016deepfool, mkadry2017towards, bai2023query, andriushchenko2020square, williams2023black} and certified robustness evaluation methods \cite{hein2017formal, weng2018evaluating, cohen2019certified, carlini2023certified}. The former generates adversarial examples through attack algorithms and measures adversarial robustness by calculating the attack success rate (ASR). However, these methods have notable limitations: on one hand, the process of generating adversarial examples is complex and resource-intensive; on the other hand, such robustness evaluation heavily depend on the selected attack methods, which may lead to biased assessments. Certified robustness employs mathematical methods to analyze the model's structure and activation functions, aiming to determine the lower bound of the perturbation that causes misclassification. However, due to its analytical nature, these methods are often difficult to implement in large and complex models, and the resulting robustness lower bounds are often overly conservative, limiting their practical applicability.

\begin{figure*}
  \centering
  \includegraphics[width=\linewidth]{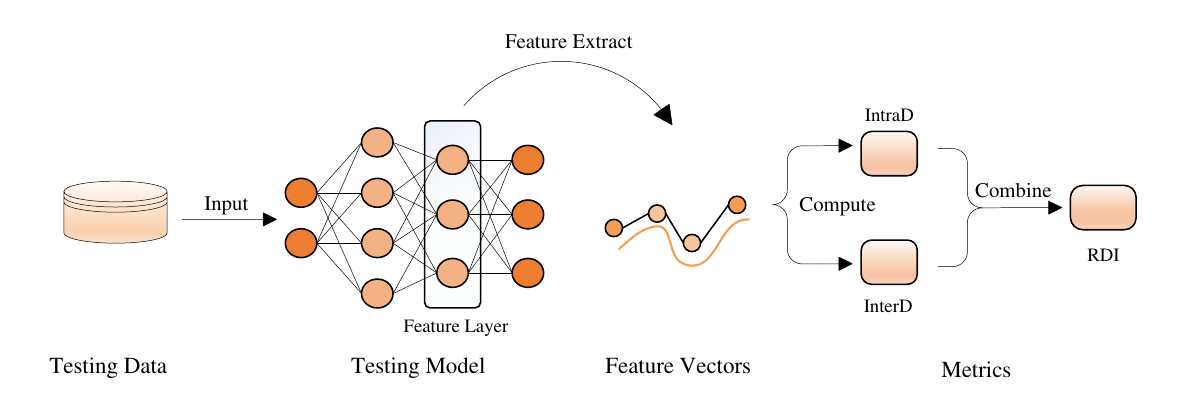}
  \caption{Framework of RDI adversarial robustness evaluation.}\label{fig:workflow}
\end{figure*}

To address the above issues, \cite{jin2022roby} proposed a robustness evaluation metric, ROBY, based on the decision boundary of the model. This metric quantifies robustness by measuring the intra-class and inter-class distances of vectors separated by the model's decision boundary. The evaluation method calculates intra-class and inter-class distances of vectors, and then combines them to produce the final adversarial robustness metric. Therefore, the accuracy of the robustness assessment depends critically on the distance calculation and the reasonableness of the weighting ratio used to combine them into the final metric. Such evaluation method brings the following issues: First, it overlooks the fact that the average magnitude of inter-class distance is larger than that of intra-class distance, resulting in difficulty to combine them to calculate the final ROBY score. Second, it is difficult to find an appropriate normalization strategy, making it challenging to compare ROBY scores across different models, which leads to a poor correlation with ASR. Finally, the metric calculation is time consuming and has a low efficiency. The reasons for the above issues are explained in Section \ref{sec:problem} of the supplementary material.

To address the limitations of existing evaluation methods, we propose a new adversarial robustness evaluation metric for DNNs,  Robustness Difference Index (RDI). RDI extracts feature vectors from the model’s feature layers and analyzes the intra-class and inter-class distances of feature vectors separated by the model's decision boundary to quantify the model's robustness. Different from ROBY, RDI adopts a novel approach to calculate intra-class and inter-class distances, as well as the final robustness evaluation metric, ensuring a more reasonable combination of these two distance indicators in the final RDI. RDI does not need a normalization strategy, allowing for effective comparison of RDI values across different models and providing an accurate assessment of adversarial robustness among different models. Moreover, RDI has high computational efficiency.

RDI doesn't rely on adversarial examples or attack algorithms, making it an metric independent of attacks. Experiments demonstrate that, across various advanced DNNs (AlexNet \cite{krizhevsky2012imagene}, ResNet50 \cite{he2016deep}, ResNet101 \cite{he2016deep}, DenseNet121 \cite{huang2017densely}, DenseNet161 \cite{huang2017densely}, MobileNetV2 \cite{howard2018inverted}, ViT \cite{dosovitskiy2020image}), RDI shows a significant correlation with the robustness gold standard metric ASR in adversarial robustness evaluation. In addition, in terms of computational efficiency, RDI outperforms both attack-based methods and ROBY. The efficiency of attack-based methods is influenced by the model architecture and attack parameter settings, while ROBY's efficiency depends on the number of classes in the dataset. However, RDI is almost unaffected by these factors. For example, the average computation time of RDI across all models on five image classification datasets is only 1/30 of the evaluation method based on PGD attack \cite{mkadry2017towards}, and for the Tiny-ImageNet dataset \cite{Deng2009ImageNet} with 200 classes, RDI's computation time is only 1/25 of ROBY's. To validate the generalizability of RDI, we conducted experiments on the speech recognition dataset SPEECHCOMMANDS \cite{warden2018speech}. The results demonstrate that this metric also exhibits excellent applicability across classification datasets of different modalities.

In summary, we aim to make the following contributions: 
\begin{itemize}
  \item We propose a model-statistical-features-based adversarial robustness evaluation metric, RDI. It is efficient, lightweight, attack-independent, and applicable to nearly all classification models.
  \item We use a novel approach to calculate intra-class and inter-class distances of feature vectors, and design the final robustness evaluation metric based on this. This approach ensures a reasonable combination of the two distance metrics to calculate RDI.
  \item We compare the RDI values across different models, as well as between natural models and adversarially trained models, and verify their strong correlation with ASR. Moreover, compared to attack-based evaluation methods and ROBY, RDI significantly improves computational efficiency.
\end{itemize}

\section{Related Works}
Currently, adversarial robustness evaluation methods can be broadly categorized into two types: attack-based evaluation methods and certified robustness evaluation methods. Additionally, decision-boundary-based evaluation methods are also worth attention.

\subsection{Adversarial Attack}
Adversarial attacks are the primary method for assessing the robustness of DNNs. Many works proposed attack algorithms aiming at generating $l_2$
or $l_{\infty}$ norm constraining adversarial perturbations. In the white-box scenario, attackers have full access to the target model's internal information, including the network architecture and gradient. As a result, many works \cite{moosavi2016deepfool, carlini2017towards, mkadry2017towards, tramer2018ensemble, wang2022iwa, liu2023gradient, gong2024cross} leveraged gradient information from the model's loss function to generate adversarial examples. In contrast, in black-box scenario, attackers have limited access to the target model, only able to obtain the output probability distribution. In this case, although direct access to gradient information is unavailable, some studies \cite{chen2017zoo, ilyas2018black, su2019one, zhou2022adversarial, bai2023query, williams2023black, wang2024adba} estimated the gradient of the loss function indirectly by querying the model’s output and then used gradient-based optimization methods to create adversarial examples. Additionally, some black-box attack methods \cite{andriushchenko2020square, guo2019simple, hong2024certifiable} do not require gradient estimation but instead optimize through specific random strategies to attack the model.

\subsection{Certified Robustness Evaluation}
Certified robustness evaluation methods primarily rely on the structure of DNNs. Through carefully designed metrics or model certifiers, these methods provide provable lower bounds on the adversarial distance required to cause misclassification. 

\cite{szegedy2013intriguing} showed that DNNs' input-output mapping often exhibits discontinuity by calculating the global Lipschitz constant for each layer, offering an initial explanation of DNN robustness. Building on this, \cite{hein2017formal} provided a lower bound for model robustness using local Lipschitz continuity conditions. Furthermore, \cite{weng2018evaluating} introduced CLEVER, a metric to evaluate the minimum perturbation needed for effective adversarial examples. \cite{cohen2019certified} provided verifiable adversarial robustness guarantees for classifiers through the method of random smoothing. \cite{carlini2023certified} proposed a more efficient boundary algorithm by integrating pre-trained denoising diffusion models with high-precision classifiers. \cite{hammoudeh2024provable} applied certified robustness to adversarial defense.

\subsection{Evaluation Methods Based On model decision boundary}
Besides above two types of robustness evaluation methods, a few studies have shown that the decision boundary is closely related to a model’s classification performance on adversarial examples and is a key factor in assessing model robustness \cite{ustun2019actionable}. Based on this finding, nearly all decision-boundary-based methods focus on exploring the relationship between the model's decision boundary and adversarial robustness. 

\cite{yousefzadeh2019investigating} calculated precise points on the model's decision boundary and provided tools to study the surface defining the boundary. \cite{he2018decision} further investigated the decision boundary characteristics of model for both adversarial and natural inputs. \cite{yang2020boundary} introduced the concept of boundary thickness for evaluating model robustness. However, these decision-boundary-based methods still rely on the generation of adversarial examples. To address this issue, \cite{jin2022roby} proposed a robustness evaluation metric, ROBY, which does not depend on adversarial examples or attack algorithms.

\begin{figure*}
  \centering
  \includegraphics[width=\linewidth]{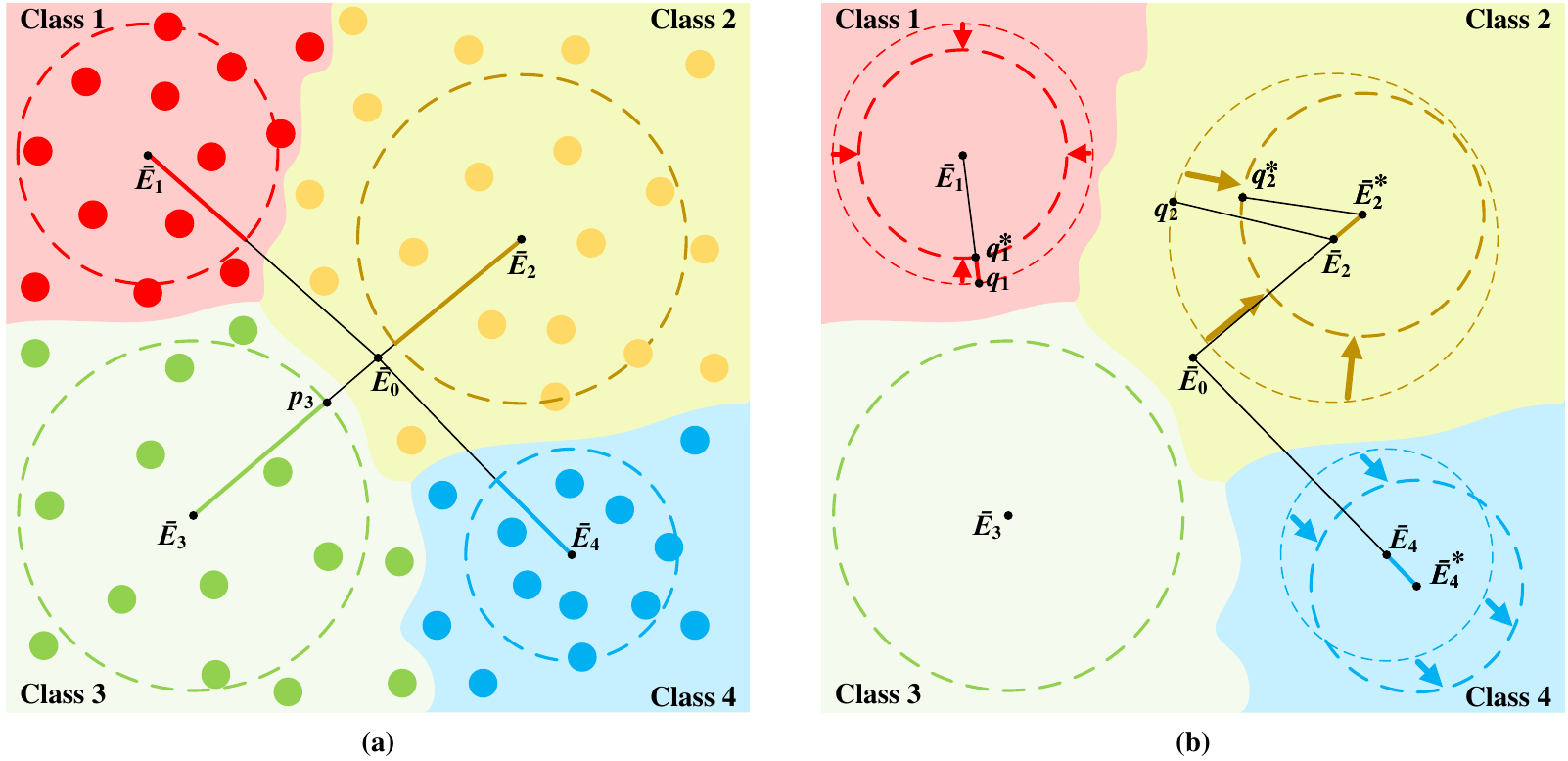}
  \caption{RDI principle schematic.}\label{fig:RDI}
\end{figure*}

\section{Proposed Evaluation Method For Adversarial robustness}\label{sec:method}
\subsection{Overview}
The framework of the proposed metric RDI is shown in Figure~\ref{fig:workflow}. First, the testing data is input into the model to be evaluated, and the features of the model's embedding space are extracted to form feature vectors. Then, for each class label of the data, compute the intra-class distance (IntraD) and inter-class distance (InterD) of the feature vectors. Finally, based on the IntraD and InterD, the adversarial robustness evaluation metric RDI is obtained. 

Our method transforms the problem of calculating the distance from test data to the decision boundary in the model's embedding space into the computation of intra-class and inter-class distances of the feature vectors. Specifically, the smaller intra-class distance and the larger inter-class distance, the stronger the model's resistance to adversarial attacks. Figure~\ref{fig:RDI} illustrates the principles behind the RDI calculation. Figure~\ref{fig:RDI}(a) presents a two-dimensional visualization of the model's decision space, where solid circles in different colors represent samples from various classes, and the boundary regions indicate the model's decision boundaries. $\bar{E}_k(k \in [1,4])$ represents the center of the feature vectors for each class, while $\bar{E}_0$ denotes the center of all sample feature vectors. $\bar{E}_k$ and $\bar{E}_0$ are computed in subsections\ref{subsec:IntraD} and \ref{subsec:InterD}, respectively.  Figure~\ref{fig:RDI}(b) demonstrates how the model’s robustness affects the RDI value. More details will be provided in the subsequent sections.

\subsection{Feature Vector Extraction}
Since the feature (penultimate) layer of the model can quickly capture and contain sufficient feature information, we use its output as the feature vectors representation of the testing data in the model's embedding space. Therefore, the dimensionality of each feature vector equals the number of model classes. For example, for an input sample $x$, we use a $K$-dimensional embedding $E(x):(e_1,...,e_K)$ as its vector representation in the model's embedding space. Suppose the model classifies sample $x$ as $k \in {R}^K$, we denote the feature vector of this sample as $E_k(x_i)$.

\subsection{Distance Function}
We use the Euclidean distance ($l_2$ norm distance) to represent the distance between two $K$-dimensional vectors $E(x_i):(e_{i1},...,e_{iK})$ and $E(x_j):(e_{j1},...,e_{jK})$ in the model's embedding space, denoted as $||E(x_i), E(x_j)||_2$:
\begin{equation}\label{eq:distance}
  ||E(x_i), E(x_j)||_2 = (\sum_{k=1}^K(e_{ik}-e_{jk})^2)^{1/2}.
\end{equation}

\subsection{IntraD Metric}\label{subsec:IntraD}
The intra-class distance (IntraD) of the feature vectors denotes the compactness of samples classified by the model as the same class. 

Since the feature vector center of each class reflects the overall characteristics of that class, we first define the center to facilitate the subsequent calculation of IntraD. For all samples classified by the model as class $k$ (denoted as set $N_k$), the mean feature vectors of this class in the model's embedding space is defined as the center of the class's feature vectors, denoted as vector $\bar{E}_k$, e.g., $\bar{E}_k$ ($k \in [1,4]$) in Figure~\ref{fig:RDI}(a). The radius of each dashed circle in the figure represents the average intra-class distance of all sample feature vectors within that class.
\begin{equation}\label{eq:eck}
 \bar{E}_k = \frac{1}{|N_k|}\sum_{x_i \in N_k}E_k(x_i).
\end{equation}
where $|N_k|$ is the total number of samples classified as class $k$ by the model, $x_i$ is the $i$-th sample in $N_k$, and $E_k(x_i)$ denotes the feature vector of sample $x_i$ classified as class $k$ by the model. The IntraD metric is the average distance of the feature vectors of samples within the same class to the center of the class's feature vectors:
\begin{equation}\label{eq:intradk}
  IntraD_k = \frac{1}{|N_k|}\sum_{i=1}^{|N_k|}||E_k(x_i), \bar{E}_k||_2.
\end{equation}
\begin{equation}\label{eq:intrad}
  IntraD = \frac{1}{K}\sum_{i=1}^{K}IntraD_k.
\end{equation}
where $IntraD_k$ represents the intra-class distance of the feature vectors for class $k$, e.g., the distance between $\bar{E}_3$ and $p_3$ in Figure~\ref{fig:RDI}(a), and $K$ is the total number of classes. Equations~\eqref{eq:intradk} and~\eqref{eq:intrad} ensure that IntraD is always positive.

\subsection{InterD Metric}\label{subsec:InterD}
The inter-class distance (InterD) of the feature vectors denotes the separation among samples classified by the model as different classes in the model's embedding space. 

Since the feature vector centers of all classes reflect the overall distribution features of the samples, to calculate InterD, we first compute the mean of the feature vectors centers for each class as the center of all class feature vectors, denoted as vector $\bar{E}_0$, e.g., $\bar{E}_0$ in Figure~\ref{fig:RDI}(a).
\begin{equation}\label{eq:ec}
  \bar{E}_0 = \frac{1}{K}\sum_{k=1}^{K}\bar{E}_k.
\end{equation}
The InterD metric is the average distance from each class's feature vectors center to the center of all class feature vectors, e.g., the average distance between $\bar{E}_0$ and $\bar{E}_k$ ($k \in [1,4]$) in Figure~\ref{fig:RDI}(a).
\begin{equation}\label{eq:interdd}
  InterD = \frac{1}{K}\sum_{i=1}^{K}||\bar{E}_k, \bar{E}_0||_2.
\end{equation}
Equation~\eqref{eq:interdd} ensures that InterD is always positive.

\subsection{RDI Metric}
Generally, models with higher robustness exhibit smaller intra-class distances and larger inter-class distances. Therefore, we define RDI to ensure a positive correlation between RDI and model robustness:
\begin{equation}\label{eq:rdi}
  RDI = \frac{InterD-IntraD}{\max(InterD, IntraD)}.
\end{equation}
Equation~\eqref{eq:rdi} ensures that RDI is in the range $[-1, 1]$.

A higher RDI value typically indicates better separation among samples from different classes and stronger grouping of samples within the same class in the model's embedding space. This results in each sample being farther from the decision boundary, meaning that larger adversarial perturbations are needed to shift the samples into adversarial regions. As a result, the model exhibits stronger robustness. Figure~\ref{fig:RDI}(b) visually illustrates the relationship between the RDI value and model robustness, revealing 3 cases leading to an increase in the RDI value. For class 1 in Figure~\ref{fig:RDI}(b), the inter-class distance remains unchanged while the intra-class distance decreases (from $[\bar{E}_1, q_1]$ to $[\bar{E}_1,q_1^*]$). For class 4, the intra-class distance remains unchanged while the inter-class distance increases (from $[\bar{E}_0,\bar{E}_4]$ to $[\bar{E}_0, \bar{E}_4^*]$). For class 2, both the intra-class distance decreases and the inter-class distance increases (the intra-class distance decreases from $[\bar{E}_2, q_2]$ to $[\bar{E}_2^*, q_2^*]$, and the inter-class distance increases from $[\bar{E}_0, \bar{E}_2]$ to $[\bar{E}_0, \bar{E}_2^*]$). Apparently, all the 3 cases increase the average distance from samples to the decision boundary, resulting in a higher RDI value. In other words, an increase in RDI indicates a larger average distance from the samples to the decision boundary, demonstrating greater robustness. Models with higher RDI values require larger perturbation to generate adversarial samples, demonstrating greater robustness. 

The calculation of RDI only requires natural samples, the property makes it an attack-independent metric. Note that the model evaluation uses samples with predicted label, not the true label. The reason is that the predicted labels reflect the model’s partitioning of the data, representing the decision boundary it has truly learned, while the true labels do not capture the model’s decision-making behavior. Robustness evaluation is meaningful only for models with a certain level of classification accuracy. In other words, it makes no sense to evaluate the robustness of a model with a very low classification accuracy.

The pseudocode of RDI calculation is presented in Algorithm~\ref{alg:RDI} of the supplementary material. First, feature vectors of the testing samples are extracted. Then, the feature centers for each class and for all samples are calculated, followed by the computation of intra-class and inter-class distances. Finally, the RDI metric is computed from these distances.

\begin{figure*}
  \centering
  \includegraphics[width=\linewidth]{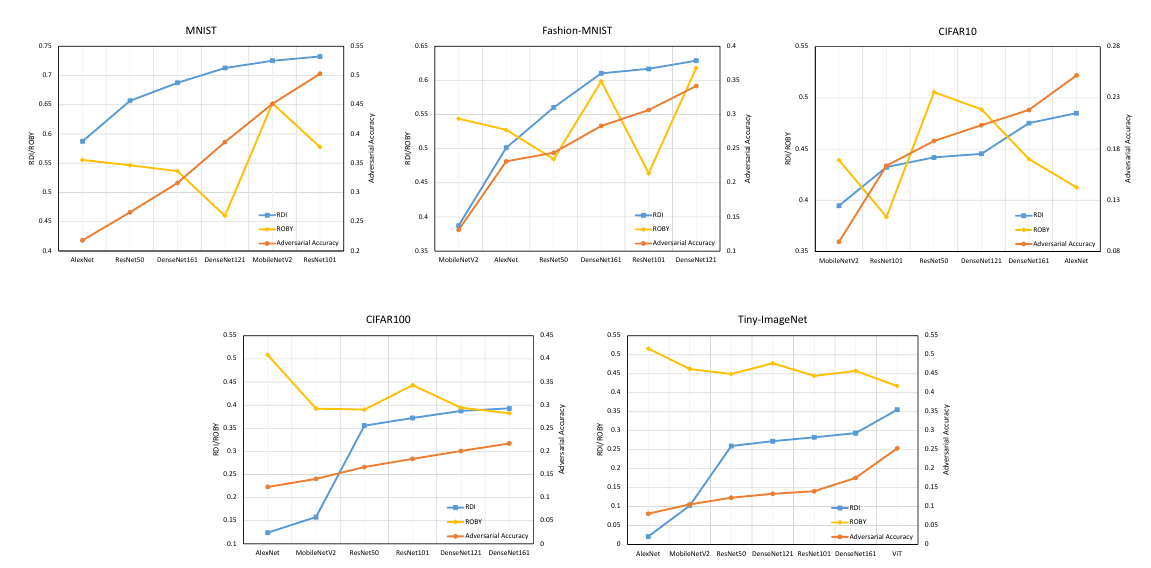}
  \caption{The relationship among RDI, ROBY, and Adversarial Accuracy across different image classification models.} \label{fig:figure3}
\end{figure*}

\section{Experiments}
We conducted four experiments to verify the effectiveness and efficiency of the RDI, focusing on the following research questions: (1) Can RDI effectively address the issues of existing evaluation methods and accurately assess adversarial robustness across different models? (2) Can RDI evaluate the robustness of adversarial trained models? (3) Does RDI demonstrate high computational efficiency? Our experiments are conducted on a server with an Intel Xeon Gold 6330 CPU, NVIDIA 4090 GPUs using PyTorch 1.13.0 on the Python 3.9.0 platform.

\begin{figure}
  \centering
  \includegraphics[width=\linewidth]{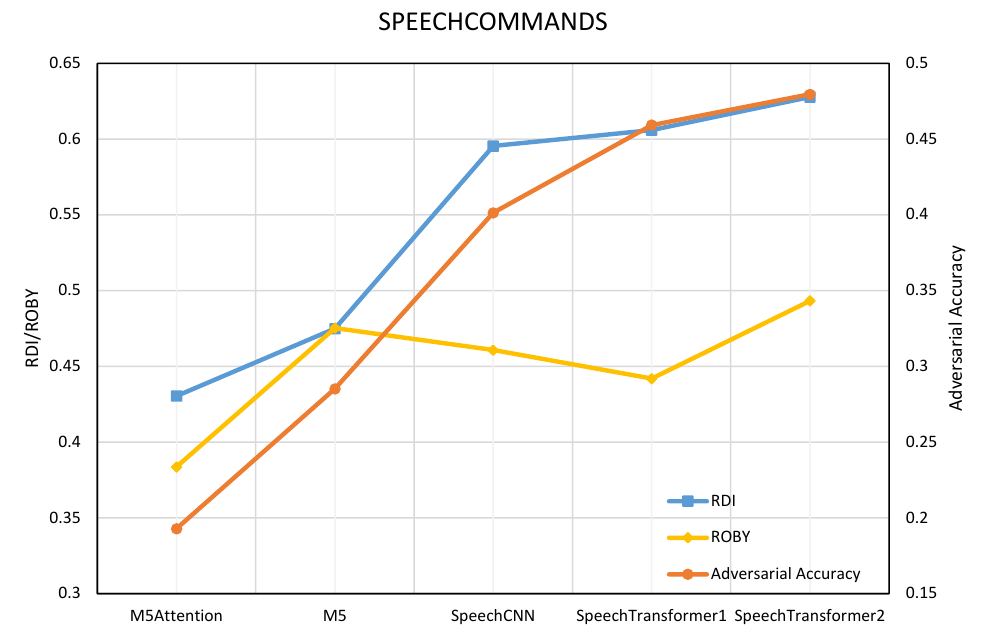}
  \caption{The relationship among RDI, ROBY, and Adversarial Accuracy across different speech recognition models.} \label{fig:figure4}
\end{figure}

\subsection{DataSets and Models}
For image classification tasks, we conducted experiments on MNIST \cite{lecun1998gradient}, Fashion-MNIST \cite{xiao2017fashion}, CIFAR10 \cite{krizhevsky2009learning}, CIFAR100 \cite{krizhevsky2009learning}, and Tiny-ImageNet \cite{Deng2009ImageNet}. For the first four datasets, we select six DNNs: AlexNet \cite{krizhevsky2012imagene}, ResNet50 \cite{he2016deep}, ResNet101 \cite{he2016deep}, DenseNet121 \cite{huang2017densely}, DenseNet161 \cite{huang2017densely}, and MobileNetV2 \cite{howard2018inverted}. For the Tiny-ImageNet dataset, to further validate the effectiveness of RDI in assessing the robustness of state-of-the-art models, we additionally included the ViT \cite{dosovitskiy2020image} model, as ViT is not well-suited for smaller datasets.

To show the broad applicability of RDI, we applied it to a speech recognition task. We conducted experiments on SPEECHCOMMANDS \cite{warden2018speech} and evaluated the robustness metrics on M5 model \cite{Arik2017Convolutional}, M5 model with an attention mechanism (M5Attention) \cite{Arik2017Convolutional}, CNN model based on DeepSpeech (SpeechCNN) \cite{abdel2014convolutional}, and two transformer-based speech recognition models (SpeechTransformer) \cite{dong2018speech}.

\subsection{Attack Methods}\label{subsec:attack}
For image classification models, we used three white-box attack methods (PGD \cite{mkadry2017towards}, RFGSM \cite{tramer2018ensemble} and C\&W \cite{carlini2017towards}) and one black-box attack method (Square Attack \cite{andriushchenko2020square}). For the MNIST and Fashion-MNIST, PGD and RFGSM ran for 40 iterations with a step size of 0.01 and perturbation size $\epsilon=0.3$, C\&W ran for 50 iterations with a step size of 0.01 and the confidence parameter $c=1$, while Square Attack ran for 5000 iterations with $\epsilon=0.3$. For the CIFAR10, PGD and RFGSM ran for 10 iterations with a step size of 0.0025 and perturbation size $\epsilon=0.1$, C\&W ran for 10 iterations with a step size of 0.01 and the confidence parameter $c=0.5$, and Square Attack ran for 200 iterations with $\epsilon=0.1$. For the CIFAR100 and Tiny-ImageNet, PGD and RFGSM ran for 10 iterations with a step size of 0.001 and perturbation size $\epsilon=0.01$, C\&W ran for 10 iterations with a step size of 0.01 and the confidence parameter $c=0.1$, while Square Attack ran for 200 iterations with $\epsilon=0.01$. For speech recognition models, we employed the adversarial audio attack method SirenAttack \cite{Du2020sirenattack}, which ran for 50 iterations with a step size of 0.05 and perturbation size $\epsilon=0.5$.

\subsection{Baseline Metric}\label{subsec:ROBY}
We used ASR, Adversarial Accuracy, and ROBY \cite{jin2022roby} as baseline comparisons with RDI. 

\textbf{ASR and Adversarial Accuracy}:
\begin{equation}\label{eq:interd}
  ASR = N_{err} / N.
\end{equation}
\begin{equation}
\label{eq:interd}
  Adversarial \ Accuracy = 1-ASR.
\end{equation}
where $N_{err}$ represents the number of classification errors on adversarial examples, and $N$ denotes the total number of adversarial samples. Since a larger RDI value indicates a more robust model, it is inversely proportional to ASR. Thus we introduced Adversarial Accuracy as one of the baseline metrics for a more intuitive comparison.

\textbf{ROBY}:
ROBY is another metric which quantifies the model's robustness via the intra-class and inter-class distances of vectors, the detailed calculation method is provided in Section \ref{sec:problem} of the supplementary material. A smaller ROBY value indicates higher model robustness.

\begin{figure*}
  \centering
  \includegraphics[width=\linewidth]{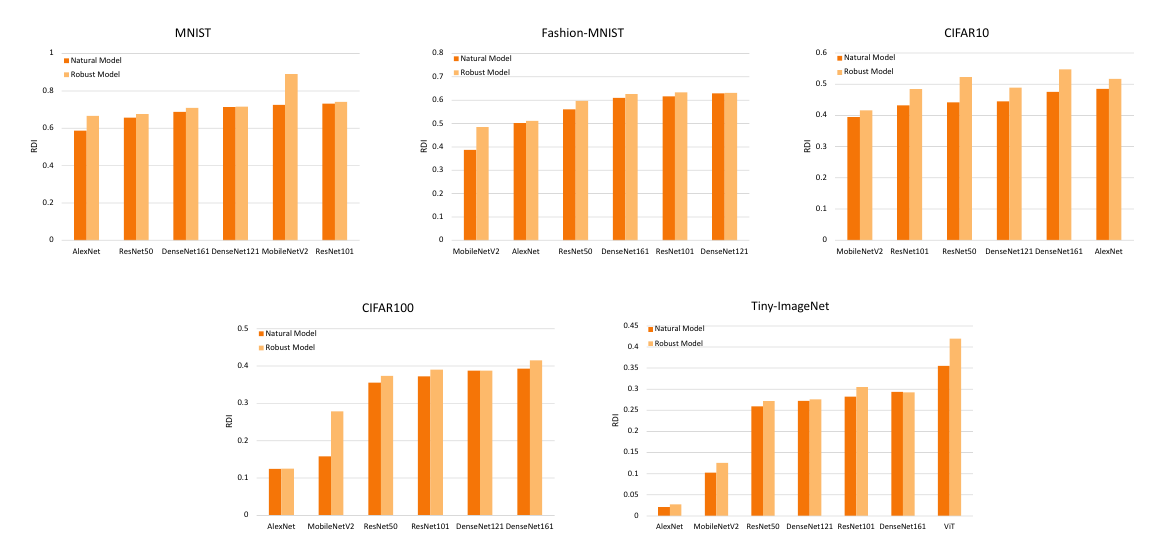}
  \caption{Comparison of RDI between natural and adversarial training models.}
  \label{fig:figure5}
\end{figure*}

\subsection{Results of RDI and Baseline Metrics For evaluating model robustness}\label{subsec:result}
To validate the precision of RDI in evaluating model adversarial robustness and highlight the issues of ROBY, we conducted experiments across multiple datasets. Specifically, we reported the average Adversarial Accuracy, ROBY, and RDI metrics for five image classification datasets under four different attack methods as well as a speech recognition dataset under SirenAttack. The relationships among these metrics are illustrated in Figure~\ref{fig:figure3} and ~\ref{fig:figure4}.

For each dataset in the figures, the models are arranged from left to right in increasing order of Adversarial Accuracy, resulting in an increasing curve for Adversarial Accuracy. Notably, RDI also forms an increasing curve. The similar trend of both metrics indicates a significant positive correlation between RDI and Adversarial Accuracy. Additionally, for more challenging datasets (such as CIFAR100 and Tiny-ImageNet), the RDI values of all models are generally lower. This phenomenon aligns with our expectations for the RDI metric, as models tend to learn ambiguous decision boundaries on challenging datasets, leading to degraded classification performance. Moreover, such ambiguous boundary result in increased intra-class distances and reduced inter-class distances in the model’s embedding space, making the model more susceptible to attack. However, for models with reasonable classification capability, the inter-class distances is typically greater than the intra-class distances. Consequently, we did not observe any instances where the RDI values fell below zero in our experiments.

ROBY should be negatively correlated with Adversarial Accuracy. However, as shown in Figure~\ref{fig:figure3} and ~\ref{fig:figure4}, ROBY exhibits clear outliers in each dataset, and its trend does not strictly follow an inverse relationship with Adversarial Accuracy. This result indicates that ROBY exhibits limited accuracy in evaluting model robustness.

Finally, we presented a more detailed evaluation results of model robustness under each dataset in Table~\ref{tab:supp-data} and Table~\ref{tab:supp-data2} in the supplementary material. For image classification datasets, we reported the accuracy, ASR under different attacks, average ASR, average Adversarial Accuracy, ROBY, and RDI values for each model, arranged in descending order of average ASR for each dataset. For the speech recognition dataset, models are ranked by descending ASR under SirenAttack, with reported metrics including accuracy, ROBY, RDI, ASR under SirenAttack, and corresponding Adversarial Accuracy.

\begin{table}
    \centering
    \caption{The computation time of RDI and ROBY.}\label{tab:cost}
    \begin{tabular}{lm{1.4cm}<{\centering}m{1.4cm}<{\centering}m{1.4cm}<{\centering}}
      \toprule 
      \multirow{2}{*}{\bfseries Dataset} & \multirow{2}{*}{\bfseries Category} & \multicolumn{2}{c}{\bfseries Time (s)}\\
      ~ & ~ &\bfseries RDI & \bfseries ROBY\\
      \midrule 
      MNIST & 10 & 3.42 & 3.95\\
      Fashion-MNIST & 10 & 3.38 & 4.02\\
      CIFAR10 & 10 & 3.43 & 3.96\\
      CIFAR100 & 100 & 4.12 & 19.84\\
      Tiny-ImageNet & 200 & 8.82 & 220.22\\
      \bottomrule 
    \end{tabular}
\end{table}

\begin{table}
    \centering
    \caption{The computation time for evaluating model robustness using RDI metric versus PGD attack method.}\label{tab:modelcost}
    \begin{tabular}{lm{2.5cm}<{\centering}m{2.5cm}<{\centering}}
      \toprule 
      \multirow{2}{*}{\bfseries Model} & \multicolumn{2}{c}{\bfseries Time (s)}\\
      ~ &\bfseries RDI & \bfseries PGD Attack\\
      \midrule 
      AlexNet & 3.38 & 11.50\\
      MobileNetV2 & 3.05 & 65.77\\
      ResNet50 & 3.61 & 108.52\\
      DenseNet121 & 4.63 & 138.66\\
      ResNet101 & 5.36 & 162.91\\
      DenseNet161 & 6.57 & 217.96\\
      Avg & 4.43 & 133.39\\
      \bottomrule 
    \end{tabular}
\end{table}

\subsection{Results of RDI Metric for natural and adversarial training models}
Previous studies \cite{guo2019simple, szegedy2013intriguing} have shown that adversarial training increases the distance between the training (testing) data and the decision boundary. In our experiments, we adopted the widely adopted PGD adversarial training method \cite{mkadry2017towards} for models adversarial training. We evaluated the effectiveness of the RDI in assessing the adversarial robustness of both natural models and adversarially trained models on five image classification datasets. The experimental results are shown in Figure~\ref{fig:figure5}.

As observed, in all five datasets, almost all adversarially trained models have higher RDI values than their natural models. This result further validates the effectiveness of the robustness evaluation metric RDI. We believe that the higher adversarial robustness of adversarially trained models is due to the increased distance between samples of different classes and the decision boundary, while the distribution of samples within the same class becomes more concentrated. As a result, the model becomes less susceptible to attack. This provides a reasonable explanation for the increase in the values of RDI after adversarial training.

\subsection{Computation Time of RDI vs. ROBY and attack-based evaluation method}

This section analyzes the time overhead of RDI, ROBY, and the PGD attack-based method to assess RDI's efficiency. Table~\ref{tab:cost} shows the computational cost of RDI and ROBY metrics on different image classification datasets. The reported times represent the average computation times for different models on each dataset, covering the entire process from model inference, feature vector extraction, and final metric calculation. Table~\ref{tab:modelcost} presents the time overhead of RDI compared to the PGD attack-based evaluation method for different models in the experiment. The reported times are the average computation times of each model across different datasets. Since the ViT model is used only on the Tiny-ImageNet dataset, it is excluded from Table~\ref{tab:modelcost} to ensure fairness in the averaging process across datasets.

As shown in Table~\ref{tab:cost}, the computational efficiency of ROBY significantly decreases as the number of classes in the dataset increases, while the efficiency of RDI is nearly unaffected by changes in the number of classes. In datasets with 10 classes, the computational efficiency of ROBY and RDI is comparable; however, as the number of classes increases, the efficiency advantage of RDI becomes more pronounced. For example, for the 100-class dataset, the computational efficiency of RDI is approximately 5 times that of ROBY, and
for the 200-class dataset, RDI’s efficiency is 25 times that of
ROBY. Table~\ref{tab:modelcost} shows that the computational efficiency of attack-based methods significantly decreases as the model complexity increases. The average computation time of RDI across all models (except ViT) on five datasets is only 1/30 of that for the PGD attack-based evaluation method. Additionally, the evaluation efficiency of attack-based methods is particularly affected by attack parameters, especially the number of iterations.

The computation of RDI only requires natural samples and does not require generating adversarial examples, significantly reducing the computational time compared to attack-based evaluation methods.

\section{Conclusion And Future Work}
This paper proposes a sample-clustering-features-based adversarial robustness evaluation metric, RDI. RDI transforms the evaluation of model robustness into the calculation of intra-class and inter-class distances of sample feature vectors in the model's embedding space, completely eliminating the reliance on adversarial examples and making it an attack-independent metric. Experimental results demonstrate that RDI shows a high correlation with Adversarial Accuracy across different models and accurately reflects the adversarial robustness differences between natural models and adversarially trained models. Additionally, RDI offers significant advantages in computational efficiency.

We believe RDI has broad applicability for various classification tasks. In future research, we will explore the potential applications of RDI in more classification scenarios and investigate its value in other robustness-related problems. For example, RDI’s efficiency enables its direct integration into the model training process, allowing for effective enhancement of model robustness without relying on adversarial training methods. This approach not only significantly reduces the overhead of adversarial training but also provides an efficient and innovative solution for robustness research. Therefore, we believe that the RDI will play a significant role in both model robustness research and applications.

\begin{acknowledgements} 
    This work was supported in part by National Natural Science Foundation of China (NO.62272117). The authors express their gratitude to the reviewers for their insightful and constructive feedback on the initial version of this paper.
\end{acknowledgements}

\bibliography{uai2025-template}

\newpage

\onecolumn

\title{RDI: An adversarial robustness evaluation metric for deep neural networks based on model statistical features\\(Supplementary Material)}
\maketitle

\begingroup
\renewcommand\thefootnote{*}
\footnotetext{Corresponding author}
\endgroup

\setcounter{footnote}{0}

\appendix
\section{Specific causes of problems with ROBY}\label{sec:problem}
The formulas for ROBY are as follows:
\begin{equation*}
    FSA_k = \frac{1}{|N_k|} \text{norm}( \sum_{i=1}^{|N_k|} dist(f_{x_i,k}, f_{c_k})).
\end{equation*}
\begin{equation*}
    FSD_{k,k+1} = \text{norm}(dist(f_{c_k}, f_{c_{k+1}})).
\end{equation*}
\begin{equation*}
    ROBY_{k,k+1} = FSA_k + FSA_{k+1} - FSD_{k,k+1}.
\end{equation*}
\begin{equation*}
    ROBY = \frac{\sum_{i=1}^{k-1} \sum_{j=i+1}^{k} \text{norm}(ROBY_{ij})}{K(K-1)/2}.
\end{equation*}
where $f_{x_i,k}$ represents the feature vector of a sample classified as class $k$, $f_{c_k}$ denotes the center of the feature vectors for class $k$, $FSA_k$ represents the intra-class distance for class $k$, $FSD_{k, k+1}$ denotes the inter-class distance between classes $k$ and $k+1$, and $ROBY_{k, k+1}$ represents the ROBY metric value between classes $k$ and $k+1$, norm(·) is the min-max normalization function, and dist(·) represents the $l_2$ norm distance. From the above equation, it can be seen that a smaller ROBY value indicates higher
model robustness, and this metric is positively correlated
with ASR.

Based on these formulas, we analyze the three main issues with the ROBY metric. First, the intra-class distance (FSA) of feature vectors in ROBY is measured by calculating the distance from the data of the same class to its class center, reflecting the "dispersion of data from the center." In contrast, the inter-class distance (FSD) is directly computed as the distance between the centers of different class data. The inconsistency in the measurement methods of these two distances leads to an unequal weight distribution when combining them into the final ROBY metric, resulting in biased evaluation outcomes. Second, ROBY uses the normalization function norm(·) for calculating FSA, FSD, and itself, employing min-max normalization. As shown in the formulas, this normalization is performed within the independent solution space of each model, making it impossible to directly compare the normalized ROBY values across different models to assess their robustness. Moreover, finding a suitable normalization strategy for this calculation is challenging. Finally, due to the many unnecessary normalization operations, ROBY also suffers from inefficiency in terms of computational cost.

\section{The pseudocode of RDI calculation}\label{sec:alg}
\begin{algorithm}[H]
\caption{RDI Calculation}\label{alg:RDI}
\begin{algorithmic}[0]
\STATE \textbf{Input:} The number of classes $K$, the samples with $K$ classes.
\STATE \textbf{Output:} RDI value
\end{algorithmic}
\begin{algorithmic}[1]
\STATE \textbf{Initialization:}  $center\_list \gets \{\}$, $IntraD\_list \gets \{\}$
\STATE // feature vectors extract
\STATE // $feature\_vectors[i][j]$ represents the feature vector of sample $j$ classified by the model as category $i$
\STATE $feature\_vectors \gets$ model($samples$)
\FOR{$k \gets 1$ \textbf{to} $K$}
    \STATE // $N_k$ represents the set of feature vectors of all samples classified as $k$ by the model
    \STATE $N_k \gets feature\_vectors[k]$
\ENDFOR
\FOR{$k \gets 1$ \textbf{to} $K$}
    \STATE // calculate the feature centers for each class
    \STATE $\bar{E}_k \gets 0$
    \FOR{$i \gets 1$ \textbf{to} $|N_k|$}
        \STATE $\bar{E}_k \gets \bar{E}_k + N_k[i]$
    \ENDFOR
    \STATE $\bar{E}_k \gets \bar{E}_k / |N_k|$
    \STATE $center\_list[k] \gets \bar{E}_k$
    \STATE // calculate the intra-class distance of class $k$
    \STATE $IntraD_k \gets 0$
    \FOR{$i \gets 1$ \textbf{to} $|N_k|$}
        \STATE $IntraD_k \gets IntraD_k + ||N_k[i], \bar{E}_k||_2$
    \ENDFOR
    \STATE $IntraD\_list[k] \gets IntraD_k / |N_k|$
\ENDFOR
\STATE // calculate the overall inter-class distance
\STATE $IntraD \gets mean(IntraD\_list)$
\STATE // calculate the overall feature center for all samples
\STATE $\bar{E}_0 \gets mean(center\_list)$
\STATE // calculate the inter-class distance
\STATE $InterD \gets 0$
\FOR{$k \gets 1$ \textbf{to} $K$}
    \STATE $InterD \gets InterD + ||center\_list[k], \bar{E}_0||_2$
\ENDFOR

\STATE $InterD \gets InterD/K$
\STATE // calculate RDI
\STATE $RDI \gets (InterD-IntraD)/\max(InterD,IntraD)$
\STATE \textbf{return} $RDI$
\end{algorithmic}
\end{algorithm}

\section{Evaluation results of model robustness}

\begin{table}[H]
    \centering
    \caption{Robustness evaluation of image classification models based on ASR, Adversarial Accuracy (AA), RDI and ROBY.}\label{tab:supp-data}
    \begin{tabular}{m{1.3cm}<{\centering}m{1.7cm}<{\centering}m{0.9cm}<{\centering}m{0.9cm}<{\centering}m{0.9cm}<{\centering}m{1.2cm}<{\centering}m{1.2cm}<{\centering}m{1.2cm}<{\centering}m{1.2cm}<{\centering}m{1.2cm}<{\centering}m{1cm}<{\centering}}
      \toprule 
      \bfseries Dataset & \bfseries model & \bfseries ACC & \bfseries ROBY & \bfseries RDI & \bfseries ASR (PGD) & \bfseries ASR (RFGSM) & \bfseries ASR (Square Attack) & \bfseries ASR (C\&W) & \bfseries ASR (avg) & \bfseries AA (avg)\\
      \midrule 
      \multirow{6}{*}{MNIST} & AlexNet & 0.9916 & 0.5555 & 0.5874 & 0.7981 & 0.8064 & 0.8121 & 0.7108 & 0.7819 & 0.2181\\
      & ResNet50 & 0.9895 & 0.5464 & 0.6566 & 0.7408 & 0.7461 & 0.7502 & 0.6988 & 0.7340 & 0.2660\\ 
      & DenseNet161 & 0.9906 & 0.5365 & 0.6873 & 0.6931 & 0.6967 & 0.7071 & 0.6382 & 0.6838 & 0.3162\\ 
      & DenseNet121 & 0.9911 & 0.4598 & 0.7127 & 0.6379 & 0.6447 & 0.6431 & 0.5304 & 0.6140 & 0.3860\\
      & MobileNetV2 & 0.9851 & 0.6523 & 0.7251 & 0.5621 & 0.5738 & 0.5813 & 0.4768 & 0.5485 & 0.4515\\
      & ResNet101 & 0.9868 & 0.5775 & 0.7324 & 0.5116 & 0.5250 & 0.5247 & 0.4274 & 0.4972 &0.5028 \\
      \midrule 
      \multirow{6}{*}{\makecell{Fashion-\\MNIST}} & MobileNetV2 & 0.9124 & 0.5439 & 0.3872 & 0.9026 & 0.9146 & 0.9098 & 0.7491 & 0.8690 & 0.1310\\
      & AlexNet & 0.9125 & 0.5275 & 0.5014 & 0.7966 & 0.8092 & 0.8043 & 0.6642 & 0.7686 & 0.2314\\ 
      & ResNet50 & 0.9121 & 0.4846 & 0.5603 & 0.7743 & 0.7841 & 0.7768 & 0.6886 & 0.7560 &0.2440 \\ 
      & DenseNet161 & 0.8877 & 0.5987 & 0.6104 & 0.7362 & 0.7413 & 0.7388 & 0.6503 & 0.7167 & 0.2833 \\
      & ResNet101 & 0.9123 & 0.4640 & 0.6169 & 0.7081 & 0.7167 & 0.7227 & 0.6258 & 0.6933 & 0.3067 \\
      & DenseNet121 & 0.9040 & 0.6183 & 0.6289 & 0.6713 & 0.6803 & 0.6894 & 0.5923 & 0.6583 & 0.3417 \\
      \midrule 
      \multirow{6}{*}{CIFAR10} & MobileNetV2 & 0.8134 & 0.4392 & 0.3947 & 0.9064 & 0.9072 & 0.9224 & 0.9063 & 0.9106 & 0.0894 \\
      & ResNet101 & 0.8340 & 0.3837 & 0.4324 & 0.8208 & 0.8237 & 0.8673 & 0.8330 & 0.8362 & 0.1638 \\ 
      & ResNet50 & 0.8422 & 0.5056 & 0.4419 & 0.7974 & 0.7986 & 0.8394 & 0.8131 & 0.8121 & 0.1879 \\ 
      & DenseNet121 & 0.8401 & 0.4888 & 0.4454 & 0.7832 & 0.7856 & 0.8286 & 0.7895 & 0.7967 & 0.2033\\
      & DenseNet161 & 0.8214 & 0.4402 & 0.4754 & 0.7622 & 0.7652 & 0.8301 & 0.7695 & 0.7818 & 0.2182\\
      & AlexNet & 0.8083 & 0.4126 & 0.4851 & 0.7270 & 0.7299 & 0.7879 & 0.7471 & 0.7480 & 0.2520 \\
      \midrule 
      \multirow{6}{*}{CIFAR100} & AlexNet & 0.6545 & 0.5084 & 0.1242 & 0.9254 & 0.9305 & 0.8487 & 0.8024 & 0.8768 & 0.1232 \\
      & MobileNetV2 & 0.6651 & 0.3925 & 0.1580 & 0.8821 & 0.8890 & 0.8547 & 0.8115 & 0.8593 & 0.1407\\ 
      & ResNet50 & 0.6928 & 0.3904 & 0.3557 & 0.8607 & 0.8652 & 0.8282 & 0.7812 & 0.8338 & 0.1662 \\ 
      & ResNet101 & 0.7032 & 0.4433 & 0.3722 & 0.8392 & 0.8462 & 0.8114 & 0.7681 & 0.8162 & 0.1838 \\
      & DenseNet121 & 0.7096 & 0.3943 & 0.3876 & 0.8206 & 0.8281 & 0.7977 & 0.7498 & 0.7991 & 0.2009 \\
      & DenseNet161 & 0.7362 & 0.3825 & 0.3931 & 0.8072 & 0.8176 & 0.7773 & 0.7294 & 0.7829 & 0.2171 \\
      \midrule 
      \multirow{7}{*}{\makecell{Tiny-\\ImageNet}} & AlexNet & 0.6452 & 0.5163 & 0.0211 & 0.9345 & 0.9477 & 0.8983 & 0.8949 & 0.9189 & 0.0811\\
      & MobileNetV2 & 0.6555 & 0.4627 & 0.1025 & 0.9145 & 0.9299 & 0.8564 & 0.8773 & 0.8945 & 0.1055\\ 
      & ResNet50 & 0.6920 & 0.4493 & 0.2595 & 0.8882 & 0.9016 & 0.8562 & 0.8616 & 0.8769 & 0.1231\\ 
      & DenseNet121 & 0.6810 & 0.4778 & 0.2723 & 0.8762 & 0.8879 & 0.8516 & 0.8501 & 0.8665 & 0.1335\\
      & ResNet101 & 0.7277 & 0.4445 & 0.2826 & 0.8728 & 0.8838 & 0.8420 & 0.8403 & 0.8597 & 0.1403\\
      & DenseNet161 & 0.7418 & 0.4576 & 0.2936 & 0.8415 & 0.8571 & 0.7979 & 0.8016 & 0.8245 & 0.1755\\
      & ViT & 0.7988 & 0.4179 & 0.3553 & 0.8042 & 0.8389 & 0.6738 & 0.6689 & 0.7465 & 0.2535\\
      \bottomrule 
    \end{tabular}
\end{table}

\begin{table}[H]
    \centering
    \caption{Robustness evaluation of speech recognition models based on ASR, Adversarial Accuracy (AA), RDI and ROBY.}\label{tab:supp-data2}
    \begin{tabular}{m{3.3cm}<{\centering}m{3cm}<{\centering}m{1cm}<{\centering}m{1cm}<{\centering}m{1cm}<{\centering}m{2.5cm}<{\centering}m{2.5cm}<{\centering}}
      \toprule 
      \bfseries Dataset & \bfseries model & \bfseries ACC & \bfseries ROBY & \bfseries RDI & \bfseries ASR (SirenAttack) & \bfseries AA (SirenAttack)\\
      \midrule 
      \multirow{6}{*}{SPEECHCOMMANDS} & M5Attention & 0.9067 & 0.3837 & 0.4305 & 0.8071 & 0.1929\\
      & M5 & 0.9575 & 0.4753 & 0.4749 & 0.7148 & 0.2852\\ 
      & SpeechCNN & 0.9251 & 0.4608 & 0.5955 & 0.5986 & 0.4014\\ 
      & SpeechTransformer1 & 0.9116 & 0.4419 & 0.6060 & 0.5408 & 0.4592 \\
      & SpeechTransformer2 & 0.9028 & 0.4932 & 0.6278 & 0.5204 & 0.4796\\
      \bottomrule 
    \end{tabular}
\end{table}

\end{document}